\documentclass[runningheads]{llncs}
\usepackage[T1]{fontenc}
\usepackage{graphicx}
\usepackage{makecell}
\usepackage{booktabs}
\usepackage{subcaption}
\usepackage{colortbl}
\usepackage{amsfonts} 
\usepackage{xcolor}
\usepackage{pifont}
\usepackage{amsmath}
\usepackage[pagebackref,breaklinks,colorlinks]{hyperref}
\usepackage{orcidlink}
\usepackage{marvosym}
\usepackage{tikz}
\usetikzlibrary{
  positioning,
  calc,
  fit,
  backgrounds,
  arrows.meta,
  shadows.blur,
  decorations.markings,
  decorations.pathreplacing
}

\definecolor{DarkGreen}{RGB}{34, 139, 34}
\begin{document}
%
\title{Can Experts Adapt Without Training? On Test-Time Modality Generalization in MVLMs}

\titlerunning{Test-Time Modality Generalization}
%


\author{Raza Imam$^{*,1}$(\Letter)\orcidlink{0000-0003-2007-6267} \and
Darakshan Rashid$^{*,3}$\orcidlink{0009-0003-1186-1017} \and
Yutong Xie$^{1}$\orcidlink{0000-0002-6644-1250} \and
Dwarikanath Mahapatra$^{2}$\orcidlink{0000-0001-9749-7858} \and
Brejesh Lall$^{3}$\orcidlink{0000-0003-2677-3071} \and
Mohammad Yaqub$^{1}$\orcidlink{0000-0001-6896-1105}}
\authorrunning{Imam et al.}
\institute{
Mohamed bin Zayed University of Artificial Intelligence, Abu Dhabi, UAE \and
Khalifa University, Abu Dhabi, UAE \and
Indian Institute of Technology Delhi, India\\
}
  
\maketitle              
\renewcommand{\thefootnote}{}%
\footnotetext{
Dataset and Code is available at: \href{https://github.com/BioMedIA-MBZUAI/MoBE-A-Test-Time-Modality-Generalization-Method}{Github}
}%
\footnotetext{
Corresponding Author: Raza Imam \Letter (\texttt{raza.imam}@mbzuai.ac.ae)
}%
\footnotetext{
Equal Contribution * \hfill Accepted to MICCAI 2026
}%
\renewcommand{\thefootnote}{\arabic{footnote}}
\begin{abstract}

Medical vision-language models (MVLMs) promise broad zero-shot generalization, yet their reliability collapses when confronted with unseen modalities and domains, precisely where clinical robustness matters most. 
To address this gap, we revisit test-time modality generalization from the perspective of Mixture-of-Experts (MoE) and ask: \textit{can experts route-and-adapt without any optimization during inference}? We identify a fundamental \textit{specialization-generalization dilemma} at test time, where blindly aggregating modality experts dilutes modality-specific knowledge, while selecting one highly confident expert risks mismatch under shift. To address this, we propose MoBE: a fully optimization-free framework that performs dynamic expert selection and adaptation at test time. MoBE combines entropy-guided dynamic routing in MoE settings with expert-wise Bayesian adaptation, enabling experts to update their confidence and adapt online without gradient updates. 
Without parametric updates, MoBE augments a static MVLM with test-time routing and online statistics, achieving average accuracy gains of +4.72, +7.17, and +4.3 over state-of-the-art TTA methods across seen, unseen, and heterogeneous medical benchmarks, highlighting the effectiveness of training-free expert adaptation for robust modality generalization.


\end{abstract}

\section{Introduction}

Recent medical vision-language models (MVLMs) enable strong zero-shot transfer across imaging tasks and modalities, but this performance often assumes test data matches training conditions \cite{ju2025delving,imam2025robustness}. In clinical deployment, distribution shifts from unseen modalities, scanners, and acquisition protocols are common, and accuracy can degrade substantially \cite{cheng2025understanding,korevaar2023failure,khan2024guardian,hanif2024baple}.

\begin{figure}[t]
    \centering
    \begin{subfigure}[t]{0.32\linewidth}
        \centering
        \includegraphics[width=\linewidth]{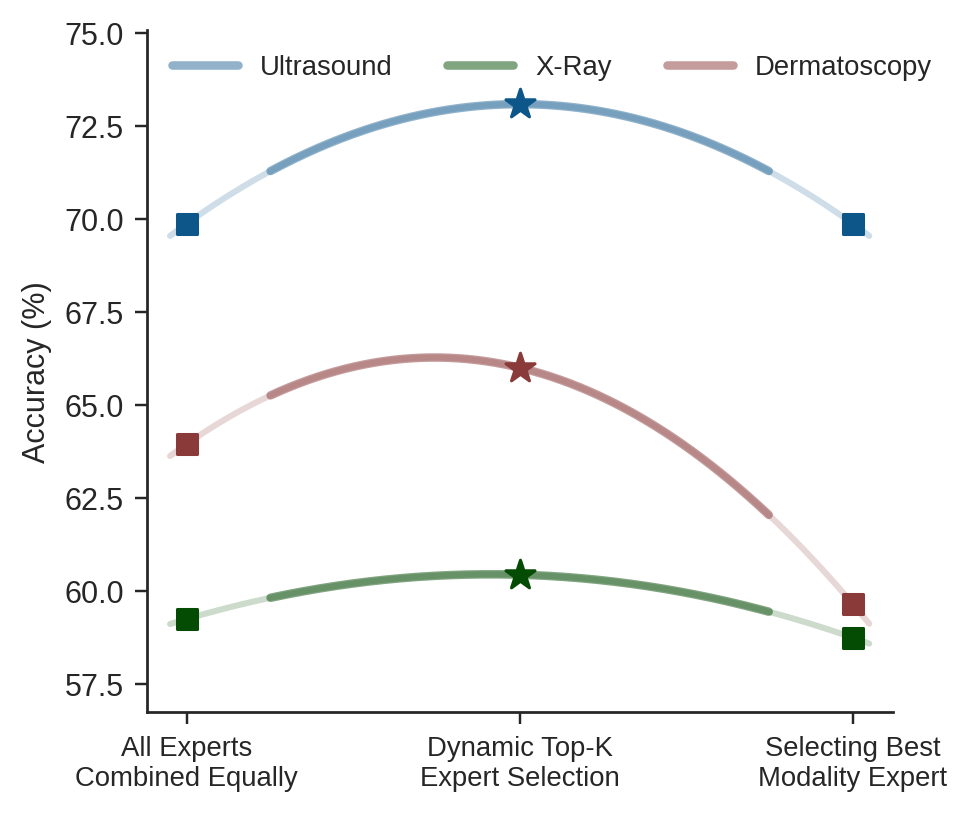}
        \caption{\small Identified Dilemma}
        \label{fig:dilemma_problem}
    \end{subfigure}
    \hfill
    \begin{subfigure}[t]{0.32\linewidth}
        \centering
        \includegraphics[width=\linewidth]{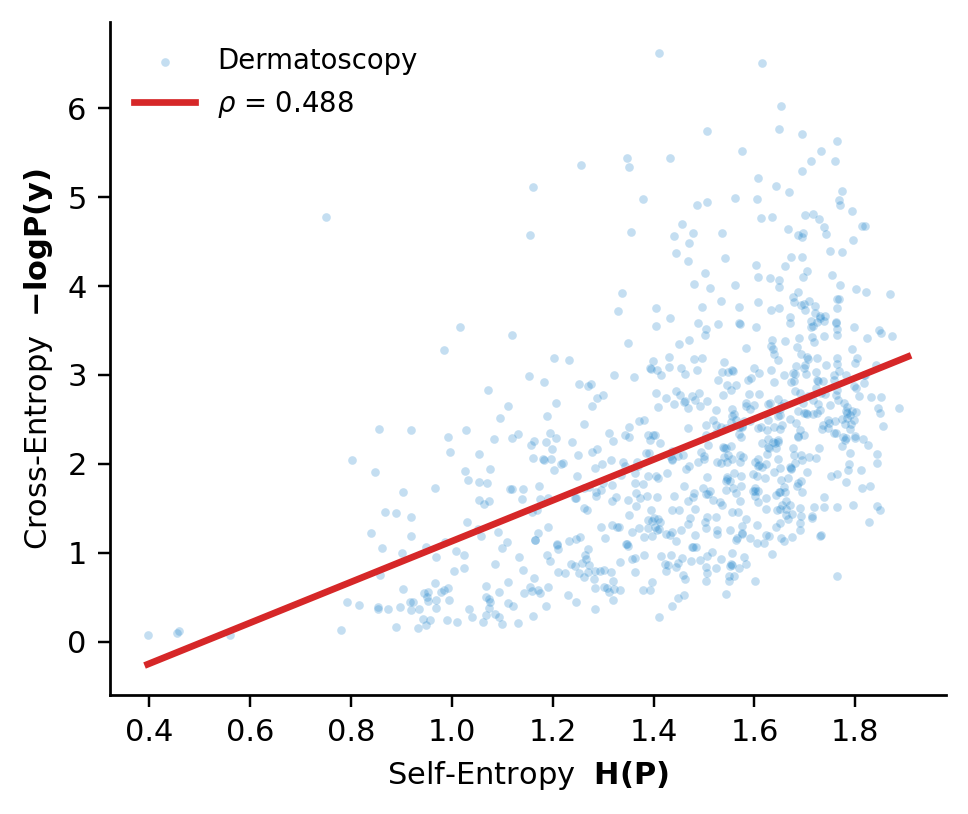}
        \caption{\small Pearson Correlation}
        \label{fig:entropy_v_oracle}
    \end{subfigure}
    \hfill
    \begin{subfigure}[t]{0.32\linewidth}
        \centering
        \includegraphics[width=0.90\linewidth]{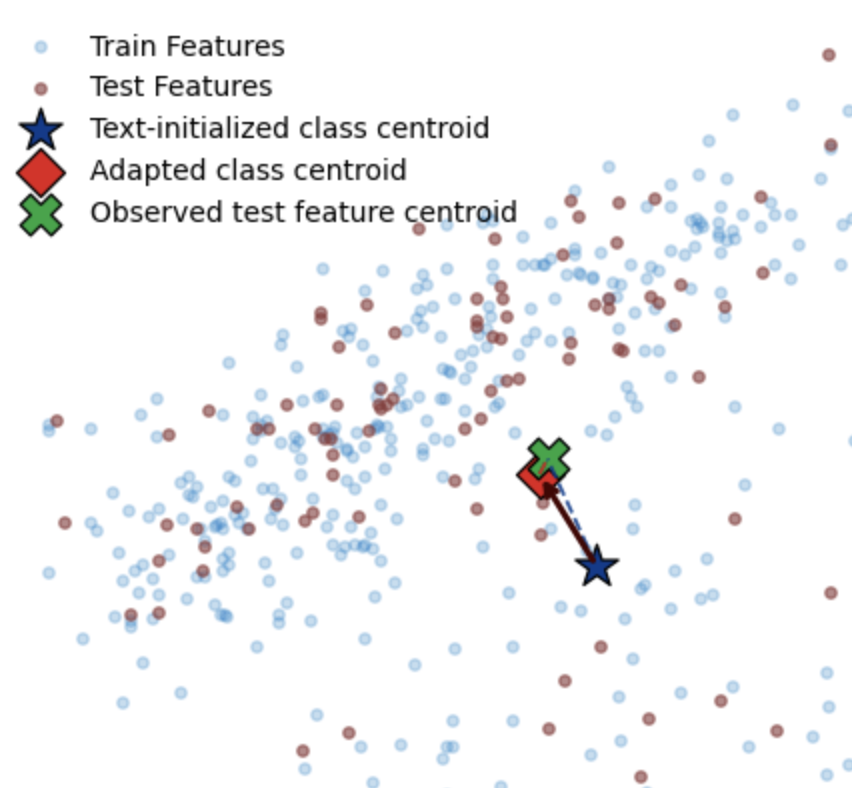}
        \caption{\small Centroid Misalignment}
        \label{fig:eba_motiv}
    \end{subfigure}
    \caption{\small (a) The specialization-generalization dilemma induced by uniform expert fusion v/s single-expert reliance at test time. (b) Routing by self-entropy closely matches oracle cross-entropy routing, as they positively correlate. (c) Despite correct routing, pretrained class centroids can misalign with test features, motivating online alignment.}
    \label{fig:concepts_overall}
\end{figure}

\noindent\textbf{Expert Specialization for Modality Heterogeneity:}
To address modality heterogeneity, prior work \cite{wang2025test} has explored expert-based architectures that partition representation learning across modalities. Mixture-of-Experts (MoE) models offer a natural mechanism for leveraging modality-specific knowledge \cite{wang2025moe}, but their effectiveness at test time critically depends on how experts are selected and combined under distribution shift. This raises a central question for test-time modality generalization: \textit{how should expert specialization be exploited when the reliability of each expert is unknown at inference?}

\noindent\textbf{Limitations of Existing Adaptation:}
MoE-based MVLMs often use fixed fusion or optimize routing at test time. For example, MoME \cite{wang2025test} adapts routing parameters via test-time optimization, aligning with gradient-based adaptation, e.g., TNT, TPT, and TTL \cite{imam2025noise,shu2022test,imam2025test}. In contrast, cache-based methods (e.g., TDA, BoostAdapter) avoid optimization via test-time memory \cite{karmanov2024efficient,zhang2024boostadapter}, but operate on a single model and do not address expert selection under modality shifts.

\noindent\textbf{The Specialization-Generalization Dilemma:}
These observations reveal an underexplored dilemma in the generalization of test-time modality: the \textit{specialization-generalization dilemma}, where blindly aggregating experts dilutes modality-specific knowledge, while committing to a single expert risks severe mismatch under shift \cite{zhong2022meta,li2022sparse}. Addressing this dilemma, therefore, requires \emph{estimating expert reliability on-the-fly at inference time}, ideally in a training-free manner that avoids backpropagation (Fig.~\ref{fig:dilemma_problem}). 
This motivates test-time adaptation in mixture-of-modality-experts settings, where dynamic expert selection and lightweight adaptation can be performed online.

\noindent\textbf{Towards Training-Free Expert Adaptation:} 
In this work, we propose an optimization-free framework that integrates dynamic expert routing with expert-wise Bayesian adaptation at test time. 
To ensure unsupervised routing matches supervised performance \cite{nikolic2025exploring,su2026variational}, Fig.~\ref{fig:entropy_v_oracle} empirically validates entropy as a cross-entropy proxy for routing. 
Also, instead of updating model parameters, each expert maintains an online posterior, accumulated across the test stream, that captures its evolving confidence and specialization, guiding both expert selection and prediction aggregation, enabling adaptation to modality shifts (Fig.~\ref{fig:eba_motiv}).

\noindent\textbf{Contributions:}
Our contributions are threefold:
    \textbf{(1)} We reveal the \emph{specialization-generalization dilemma} that cripples test-time modality adaptation in Medical VLMs;
    \textbf{(2)} We present MoBE: a fully \emph{optimization-free} inference framework uniting dynamic expert routing with per-expert Bayesian adaptation; and
    \textbf{(3)} We demonstrate \emph{consistent state-of-the-art gains} across standard and heterogeneous medical benchmarks, proving training-free expert specialization enables robust modality generalization.

\section{Methodology}
\begin{figure}[t]
    \centering
    \includegraphics[width=0.80\linewidth]{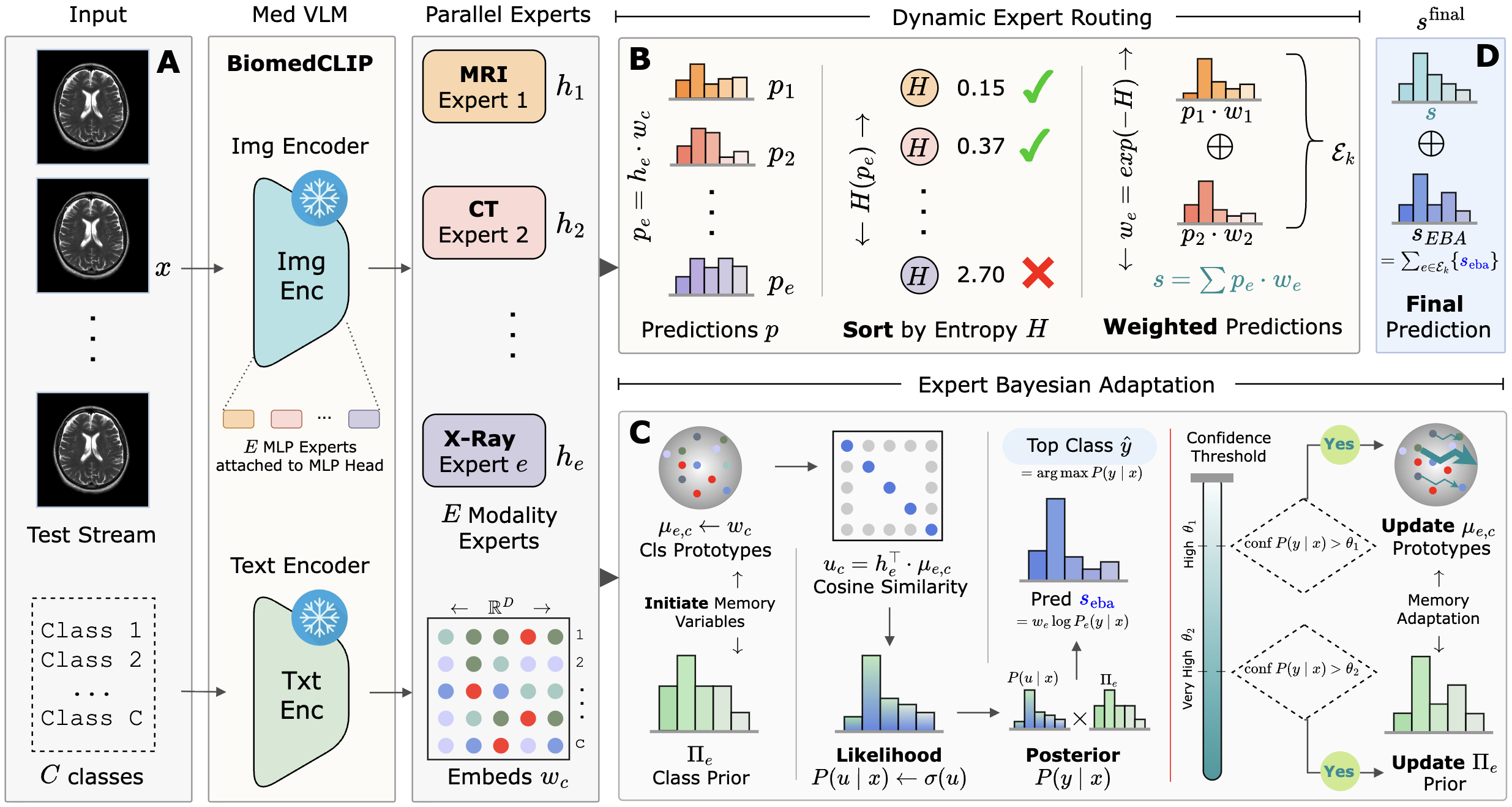}
    \caption{\small \textbf{MoBE Test-Time Adaptation.} (A) Frozen BiomedCLIP encoder processes an incoming stream of test images. (B) Dynamic-$k$ entropy routing selects the most reliable modality experts from parallel MLP heads. (C) Selected experts are inverse-entropy weighted and refined by EBA through confidence-gated prototype and prior updates using the test stream. (D) Final predictions blend routed logits with Bayesian-adapted outputs.}
    \label{fig:method}
\end{figure}

Our modality-generalizable test-time adaptation is a two-stage process: first, dynamic-$k$ routing addresses the \textit{specialization-generalization dilemma} by selecting contextually reliable modality experts through predictive uncertainty; second, Expert Bayesian Adaptation (EBA) refines the predictions of selected experts through online posterior updates via exponential moving averages (Fig.~\ref{fig:method}). This yields an optimization-free, MoE-based TTA system that continuously adapts how fixed representations are combined and refined, achieving robust performance without gradient updates or source data access.

\subsection{Preliminaries: Modality-Specialized MVLM Backbone}

Given a sequence of test images $\{x_i\}_{i=1}^n$ that arrive sequentially from an unseen modality, the model is required to predict the label for each image $x_i$ immediately in an online manner. For each incoming image, the frozen image encoder $f_\theta$ of a multimodal vision-language model (MVLM) (e.g., BiomedCLIP \cite{zhang2023biomedclip}) extracts a visual representation $h_i = f_\theta(x_i) \in \mathbb{R}^D$. Zero-shot classification is performed by computing the similarity between the visual feature $h_i$ and text-encoder embeddings $w_c \in \mathbb{R}^D$, corresponding to class prompts of the form ``a [class] of [modality]'', yielding logits $s_{i,c} = h_i^\top w_c$.

To leverage modality-specific knowledge, following MoME \cite{wang2025test}, we attach $E$ modality-specialized MLP experts $\{g_e\}_{e=1}^E$ in parallel to the frozen image encoder's classification head. Each $g_e: \mathbb{R}^D \to \mathbb{R}^C$ maps visual features to class logits tailored to modality $m_e$, producing $s_e = g_e(h)$ and softmax probabilities $p_e = \mathrm{softmax}(s_e)$. The challenge is selecting and combining these parallel experts when the test modality is unknown.


\subsection{MoBE: \underline{M}ixture \underline{o}f \underline{B}ayesian \underline{E}xperts in Medical Imaging} 
\noindent\textbf{A.~Dynamic-$k$ Entropy-Guided Expert Routing:}
As shown in Fig. \ref{fig:dilemma_problem}, both uniform expert fusion and single-expert selection suffer performance degradation by either diluting modality knowledge or risking catastrophic mismatch. We resolve this through dynamic-$k$ routing that adaptively gates experts based on relative predictive uncertainty.

For each test sample, we compute predictive entropy $H(p_e) = -\sum_c p_e(c)\log p_e(c)$ across all $E$ experts, where lower $H(p_e)$ signals higher modality alignment. Rather than fixed top-$k$, we apply relative thresholding: sort by ascending entropy $H_1 \leq H_2 \leq \dots \leq H_E$, then select $k$ experts as:
\begin{equation}
    \mathcal{E}_k = \{e_j \mid H_j - H_1 < \tau \}, \quad k = |\mathcal{E}_k|, \quad j \in {1,\dots,E},
\end{equation}
where $\tau$ is a modality-agnostic gap threshold. This way, we select and gate experts who are nearly as confident as the most confident expert, and exclude clearly mismatched ones. A minimum of two experts is enforced when $E>1$, maintaining expert diversity and inference efficiency~\cite{wang2025test}.

Following observations from TTL and DeYo~\cite{imam2025test,lee2024entropy}, we note that inverse-entropy weighting produces mixing coefficients that better capture relative confidence differences than softmax:
\begin{equation}
    w_e \propto \frac{1}{\exp(H(p_e))}, \quad e \in \mathcal{E}_k.
\end{equation}
Gated logits become the weighted ensemble $s = \sum_{e\in\mathcal{E}_k} w_e s_e$, where the gating function naturally emphasizes confident modality experts.

\noindent\textbf{B.~Expert Bayesian Adaptation (EBA):}
Once top experts $\mathcal{E}_k$ are selected via dynamic routing, 
the next step is to adapt their predictions online as the test stream evolves under a distribution shift.
Motivated by statistical methods for natural images~\cite{han2024dota,zhou2025bayesian}, which frame predictions as a factorization of likelihood $P(x\mid y)$ and prior $P(y)$, we observe that existing approaches primarily adapt likelihoods while neglecting prior shifts. To address this, our {Expert Bayesian Adaptation (EBA)} module maintains per-expert Bayesian posteriors and jointly adapts both factors online, without updating any backbone or expert parameters.

For each selected expert $e \in \mathcal{E}_k$, EBA maintains two persistent \emph{memory variables} across the test stream:
(i) class prototypes $\mu_{e,c}\in\mathbb{R}^D$ (one per class $c$, serving as a moving representative feature/centroid), and
(ii) a class-prior vector $\Pi_e\in\Delta^{C-1}$ (a length-$C$ distribution capturing the expert's historical tendency to predict each class).
As additional test samples are observed, these accumulated statistics become increasingly accurate summaries of the underlying test distribution.

\textbf{\textit{Initialization:}}
Prototypes are initialized from pretrained text embeddings $\{w_c\}_{c=1}^C$ as semantically grounded class anchors:
$\mu_{e,c} \leftarrow {w_c}/{\lVert w_c\rVert_2},\text{where~} c=1,\dots,C.$
The class prior is initialized uniformly:
$
\Pi_e \leftarrow [{1}/{C},{1}/{C},\dots,{1}/{C}].
$
We also initialize per-expert, per-class update counters $c_{1,e,c}, c_{2,e,c}\in\mathbb{N}$; these count confident prototype and prior updates, respectively, and act as running-average denominators.

\textbf{\textit{Prediction:}}
Given a test feature vector $h$, we score each class by its similarity to the corresponding prototype:
$
u_c=\tau_{\mathrm{EBA}}\cdot h^\top \mu_{e,c},
$
and convert scores into a probability vector
$
P(u\mid x)=\mathrm{softmax}(u),
$
which we treat as a likelihood-like quantity indicating how well the input matches each class. To incorporate the expert's stored history to attain full \textit{posterior}, we reweight these probabilities by the prior $\Pi_e$ (elementwise) and renormalize:
\begin{equation}
\label{eq:eba_posterior_minfix}
\tilde{P}(y\mid x)=P(u\mid x)\odot \Pi_e,
\qquad
P(y\mid x)=\frac{\tilde{P}(y\mid x)}{\sum_{c}\tilde{P}(y=c\mid x)}.
\end{equation}
Intuitively, this combines \emph{current similarity to prototypes} with \emph{what this expert has been seeing recently}.

\textbf{\textit{Confidence-gated adaptation:}}
EBA updates its memory only when it is confident, using confidence-gated running averages.
(i) \textit{Prototype update (threshold $\theta_1$):}
For high-confidence predictions (i.e., if $\max_c P(y=c\mid x) > \theta_1$),
we update only the prototype of the predicted class
$\hat{y}=\arg\max P(y\mid x)$ by averaging it with the new feature $h$:
\begin{equation}
\mu_{e,\hat{y}} \leftarrow \frac{c_{1,e,\hat{y}}\,\mu_{e,\hat{y}} + h}{c_{1,e,\hat{y}} + 1},
\qquad
c_{1,e,\hat{y}} \leftarrow c_{1,e,\hat{y}} + 1,
\end{equation}
followed by re-normalizing prototypes to unit length.
(ii) \textit{Prior update (higher threshold $\theta_2$):}
When confidence exceeds a higher threshold (i.e., if $\max_c P(y=c\mid x) > \theta_2$),
we additionally update the class prior using the current posterior $P(y\mid x)$:
\begin{equation}
\Pi_e \leftarrow \frac{c_{2,e,\hat{y}}\,\Pi_e + P(y\mid x)}{c_{2,e,\hat{y}} + 1},
\qquad
c_{2,e,\hat{y}} \leftarrow c_{2,e,\hat{y}} + 1.
\end{equation}

As $c_{1,e,c}$ and $c_{2,e,c}$ increase, each update receives less weight, making $\mu_{e,c}$ and $\Pi_e$ progressively more stable.
This dual adaptation mechanism: (i) moving prototypes toward the observed test feature manifold, and (ii) adjusting class priors based on a history of highly confident posteriors; aligning to the target distribution while preserving pretrained semantic structure.
EBA is applied independently to each $e \in \mathcal{E}_k$, with aggregated logits as: $s_{\mathrm{EBA}} = \sum_{e \in \mathcal{E}_k} w_e \log P_e(y \mid x)$.
Final predictions fuse entropy-routed logits with EBA-adapted logits:
$
    s^\mathrm{final} = (1 - \lambda)\, s + \lambda\, s_\mathrm{EBA},
$
where $\lambda \in [0,1]$ controls the strength of adaptation.

\noindent\textbf{C. Training \& Test-Time Inference:}
Each modality-specific MLP $g_e$ trains independently on source data from modality $m_e$, minimizing cross-entropy using the frozen image encoder $f_\theta$. 
No routing or cross-modal mixing occurs during training.
\noindent\textit{Test-Time Inference:} The image-text encoders and experts remain frozen. Per test batch: (1) extract $h$ and per-expert logits $\{s_e\}$; (2) compute entropies $\{H(p_e)\}$ and gate dynamic-$k$ set $\mathcal{E}_k$; (3) apply inverse-entropy weighting to produce $s$; (4) update EBA posteriors for selected experts; (5) blend via $s^\mathrm{final}$. 

\section{Experiments and Results}
\subsection{Experimental Settings}

\noindent\textbf{Benchmarks:}
Following prior work on modality-level TTA~\cite{wang2025test}, we evaluate on a combined MedMNIST$+$Med-VTAB benchmark~\cite{mo2024large}, covering 11 datasets across 9 imaging modalities (e.g., X-ray, CT, pathology, microscopy, ultrasound, OCT, and MRI). Med-VTAB adds MRI to mitigate modality gaps in MedMNIST and enables controlled assessment of modality generalization. Furthermore, we evaluate on a heterogeneous few-shot suite~\cite{koleilat2025biomedcoop} with 11 datasets spanning 10 organs and 9 modalities (including MRI, ultrasound, endoscopy, and histopathology), to test robustness under more diverse clinical imaging conditions.

\noindent\textbf{Implementation Details:}
We use the pretrained BiomedCLIP ViT-B/16 backbone \cite{zhang2023biomedclip} with frozen image and text encoders. We attach $E{=}5$ modality-specialized MLP experts in parallel to the image encoder’s classification head. Similar to \cite{wang2025test}, experts are derived from ROCOV2~\cite{ruckert2024rocov2} by partitioning data into five modality groups (CT, MRI, X-ray, ultrasound, angiogram) and fine-tuning only the vision-side MLP layers per modality; the resulting MLP weights are extracted as fixed expert branches in MoBE. These modality experts have the same architecture as in BiomedCLIP's MLP. All images are processed using the standard BiomedCLIP preprocessing with no additional augmentation. We evaluate with batch size 1 on a single NVIDIA A6000 (48GB) and report Top-1 accuracy. Dataset-specific hyperparameter values for MoBE are available at \href{https://github.com/BioMedIA-MBZUAI/MoBE-A-Test-Time-Modality-Generalization-Method}{Github}.


{\renewcommand{\arraystretch}{1.1}
\begin{table}[b]
\centering
\caption{\small Performance of different VLMs and TTA methods on MedMNIST and Med-VTAB datasets \cite{wang2025test} (\% Accuracy).}
\label{tab:seen_results}
\resizebox{0.80\textwidth}{!}{%
\begin{tabular}{lccccccc}
\hline
Method &
  \makecell{ChestMNIST} &
  \makecell{MedVTAB} &
  \makecell{BreastMNIST} &
  \makecell{OrganAMNIST} &
  \makecell{OrganCMNIST} &
  \makecell{OrganSMNIST} &
  \makecell{Average} \\ 
  \hline
Modality &
  X-Ray &
  MRI &
  Ultrasound &
  CT &
  CT &
  CT & Accuracy
  \\ 
  \hline
BiomedCLIP & 53.29 & 51.77 & 32.69 & 25.47 & 17.60 & 17.15 & \cellcolor{teal!10}33.00 \\
CoOp [52]  & 53.30 & 52.71 & 33.58 & 26.04 & 17.65 & 17.59 & \cellcolor{teal!12}33.48 \\
CoCoOp [51]& 53.36 & 52.75 & 33.51 & 26.19 & 17.88 & 17.71 & \cellcolor{teal!14}33.57 \\
TENT [37]  & 41.81 & 45.86 & 30.15 & 15.47 & 9.61  & 10.02 & \cellcolor{teal!5}25.49 \\
EATA [30]  & 44.63 & 51.98 & 30.83 & 22.31 & 15.82 & 15.55 & \cellcolor{teal!8}30.19 \\
TPT [36]   & 53.33 & 53.22 & 35.77 & 26.12 & 17.67 & 17.60 & \cellcolor{teal!16}33.95 \\
WATT [31]  & 53.42 & 56.10 & 35.97 & 25.89 & 16.84 & 16.77 & \cellcolor{teal!18}34.17 \\
TDA [18]   & 53.50 & 55.69 & 45.87 & 26.40 & 17.91 & 17.82 & \cellcolor{teal!22}36.20 \\
MoME       & 54.72 & 58.12 & 52.56 & \textbf{29.19} & 18.45 & 19.12 & \cellcolor{teal!28}38.69 \\
\textbf{MoBE (Ours)} & \textbf{60.44} & \textbf{60.72} & \textbf{73.08} & 24.37 & \textbf{19.84} & \textbf{21.99} & \cellcolor{teal!38}\textbf{43.41} \\
 \hline
\end{tabular}%
}
\end{table}
}
{\renewcommand{\arraystretch}{1.1}
\begin{table}[t]
\centering
\caption{\small Performance of different VLMs and TTA methods on MedMNIST datasets with unseen modalities \cite{wang2025test} (\% Accuracy).}
\label{tab:unseen_results}
\resizebox{0.85\textwidth}{!}{%
\begin{tabular}{lcccccc}
\hline
Method &
  \makecell{PathMNIST} &
  \makecell{DermaMNIST} &
  \makecell{OCTMNIST} &
  \makecell{BloodMNIST} &
  \makecell{TissueMNIST} &
  \makecell{Average} \\ 
\hline
Modality &
  Colon Pathology &
  Dermatoscope &
  Retinal OCT &
  Microscope &
  Microscope &
  Accuracy
  \\ 
\hline
BiomedCLIP [49] & 28.31 & 17.71 & 27.70 & 23.82 & 4.89 & \cellcolor{teal!10}20.49 \\
CoOp [52]       & 28.59 & 18.07 & 29.98 & 24.30 & 4.95 & \cellcolor{teal!12}21.18 \\
CoCoOp [51]     & 28.50 & 18.15 & 30.82 & 24.36 & 5.10 & \cellcolor{teal!14}21.39 \\
TENT [37]       & 20.48 & 14.05 & 21.96 & 15.54 & 3.22 & \cellcolor{teal!5}15.05 \\
EATA [30]       & 24.72 & 14.20 & 27.82 & 19.35 & 4.60 & \cellcolor{teal!8}18.14 \\
TPT [36]        & 28.34 & 20.03 & 31.57 & 25.14 & 5.07 & \cellcolor{teal!16}22.03 \\
WATT [31]       & 28.45 & 29.64 & 33.50 & 25.12 & 4.91 & \cellcolor{teal!20}24.32 \\
TDA [18]        & 28.21 & 23.18 & 31.65 & 24.79 & 4.96 & \cellcolor{teal!22}22.56 \\
MOME            & 29.74 & 57.51 & 36.20 & \textbf{26.31} & 5.18 & \cellcolor{teal!28}30.99 \\
\textbf{MoBE}   & \textbf{48.69} & \textbf{66.00} & \textbf{54.20} & 15.70 & \textbf{6.22} & \cellcolor{teal!38}\textbf{38.16} \\
\hline
\end{tabular}%
}
\end{table}
}

{\renewcommand{\arraystretch}{1.1}
\begin{table}[t]
\centering
\caption{\small Performance of different VLMs and TTA methods on heterogeneous medical datasets \cite{koleilat2025biomedcoop} (\% Accuracy).}
\label{tab:hetereognenous_results}
\begin{subtable}{\textwidth}
\centering
\resizebox{0.85\textwidth}{!}{%
\begin{tabular}{lcccccc}
\hline
Method &
  BTMRI &
  COVID-QU-Ex &
  CTKIDNEY &
  DermaMNIST &
  Kvasir &
  CHMNIST \\ 
Modality &
  MRI &
  X-Ray &
  CT &
  Dermatoscopy &
  Endoscopy &
  Histopathology \\
\hline
BiomedCLIP [49] & 51.77 & 68.72 & 49.01 & 17.71 & 47.25 & 23.67 \\
TDA [18]        & 55.69 & 64.45 & 54.71 & 23.18 & \textbf{52.08} & \textbf{27.53} \\
\textbf{MoBE} & \textbf{62.90} & \textbf{69.86} & \textbf{55.72} & \textbf{66.00} & 50.42 & 24.87 \\
\hline
\end{tabular}%
}
\end{subtable}
\hfill
\begin{subtable}{\textwidth}
\centering
\resizebox{0.85\textwidth}{!}{%
\begin{tabular}{lcccccc}
\hline
Method &
  LC25000 &
  RETINA &
  KneeXray &
  OCTMNIST &
  BUSI &
  Average \\ 
Modality &
  Histopathology &
  Fundus Photography &
  X-Ray &
  OCT &
  Ultrasound &
  Accuracy \\
\hline
BiomedCLIP [49] & 33.87 & 23.74 & \textbf{40.28} & 27.70 & 41.10 & \cellcolor{teal!15}39.24 \\
TDA [18]        & \textbf{40.80} & 27.60 & 38.47 & 31.65 & 45.34 & \cellcolor{teal!25}42.19 \\
\textbf{MoBE} & 37.68 & \textbf{29.73} & 40.10 & \textbf{54.20} & \textbf{45.76} & \cellcolor{teal!38}\textbf{46.49} \\
\hline
\end{tabular}%
}
\end{subtable}
\end{table}
}

\subsection{Comparison with State-of-the-art}
We benchmark MoBE against prominent adaptation methods on BiomedCLIP, including TENT~\cite{wang2020tent}, EATA~\cite{niu2022efficient}, CoOp~\cite{zhou2022conditional}, CoCoOp~\cite{zhou2022learning}, TPT~\cite{shu2022test}, WATT~\cite{osowiechi2024watt}, and TDA~\cite{niu2022efficient}. While CoOp and CoCoOp are prompt-learning methods rather than strict TTA, we include them following common protocols in \cite{wang2025test}.

\noindent\textbf{Results on MedMNIST and MedVTAB (seen modalities):}
Tab.~\ref{tab:seen_results} reports datasets whose modalities are covered by our modality-specialized experts. MoBE achieves the best average accuracy (43.41\%), improving over BiomedCLIP (33.00\%) and MoME (38.69\%). Gains are consistent on ChestMNIST, MedVTAB, and BreastMNIST (notably 73.08\% on BreastMNIST), and also on OrganCMNIST/OrganSMNIST. OrganAMNIST is the main exception, likely due to weaker modality cues and heterogeneous class semantics that make entropy-based expert selection less reliable.

\noindent\textbf{Results on MedMNIST with unseen modalities:}
Tab.~\ref{tab:unseen_results} evaluates generalization to modalities unseen during expert pre-training. MoBE substantially improves the average (38.16\%) over BiomedCLIP (20.49\%), TDA (22.56\%), and MoME (30.99\%), with strong gains on PathMNIST, DermaMNIST, OCTMNIST, and TissueMNIST. BloodMNIST remains challenging, plausibly because microscopy shifts are not well covered by our five expert modalities, reducing routing discriminability.

\noindent\textbf{Results on Heterogeneous Dataset:}
Tab.~\ref{tab:hetereognenous_results} extends evaluation to heterogeneous datasets (e.g., endoscopy, fundus photography, histopathology). Since MoME is not reproducible here due to code unavailability, we compare BiomedCLIP, TDA, and MoBE. Averaged over 11 datasets and 9 modalities, MoBE achieves the best average (46.49\%) vs.\ TDA (42.19\%), with large gains on OCTMNIST and competitive results on RETINA/BUSI. TDA is slightly stronger on some sets (e.g., Kvasir/CHMNIST), where style-driven shifts may favor cache-based likelihood adaptation over expert selection.

\begin{figure}[t]
\centering
\begin{minipage}{0.40\textwidth}
\centering
{\renewcommand{\arraystretch}{1.2}
\captionof{table}{\small Performance comparison with different frozen backbones.}
\resizebox{\textwidth}{!}{%
\begin{tabular}{lccc}
\hline
Backbone $\downarrow$ & Seen-Avg & Unseen-Avg & Average \\
\hline
PubmedCLIP & 35.89 & 10.37 & \cellcolor{teal!15}23.13 \\
+\textbf{MoBE} (Ours) & \textbf{48.13} & \textbf{10.98} & \cellcolor{teal!38}\textbf{29.56} \\
\hline
BiomedCLIP & 35.45 & 23.01 & \cellcolor{teal!15}29.23 \\
+\textbf{MoBE} (Ours) & \textbf{40.14} & \textbf{57.35} & \cellcolor{teal!38}\textbf{48.74} \\
\hline
\end{tabular}%
\label{tab:backbone}
}
}
\end{minipage}
\begin{minipage}{0.59\textwidth}
\centering
{\renewcommand{\arraystretch}{1.2}
\captionof{table}{\small Computation analysis on X-ray modality (ChestMNIST, $\sim$22k samples).}
\resizebox{\columnwidth}{!}{%
\begin{tabular}{lccccc}
\hline
Method~$\downarrow$ & No Backprop? & Infer-Time & Speed$^*$ & Accuracy & $\Delta$Gain \\ 
\hline
BiomedCLIP & \textcolor{DarkGreen}{\ding{52}} & 2m30s & 144 i/s & 53.29 & 0.00 \\
MoME       & \textcolor{purple}{\ding{56}} & \cellcolor{teal!15}12m57s & \cellcolor{teal!15}28 i/s & \cellcolor{teal!15}54.72 & \cellcolor{teal!15}1.43 \\
\textbf{MoBE} (Ours)& \textcolor{DarkGreen}{\ding{52}} & \cellcolor{teal!38}\textbf{11m10s} & \cellcolor{teal!38}\textbf{35 i/s} & \cellcolor{teal!38}\textbf{60.44} & \cellcolor{teal!38}\textbf{7.15} \\ \hline
\multicolumn{6}{l}{\footnotesize{$^*$i/s denotes inference speed in images-per-second}}
\end{tabular}%
\label{tab:compute}
}
}
\end{minipage}
\end{figure}

\begin{figure}[t]
    \centering
    \begin{subfigure}[t]{0.32\linewidth}
        \centering
        \includegraphics[width=\linewidth]{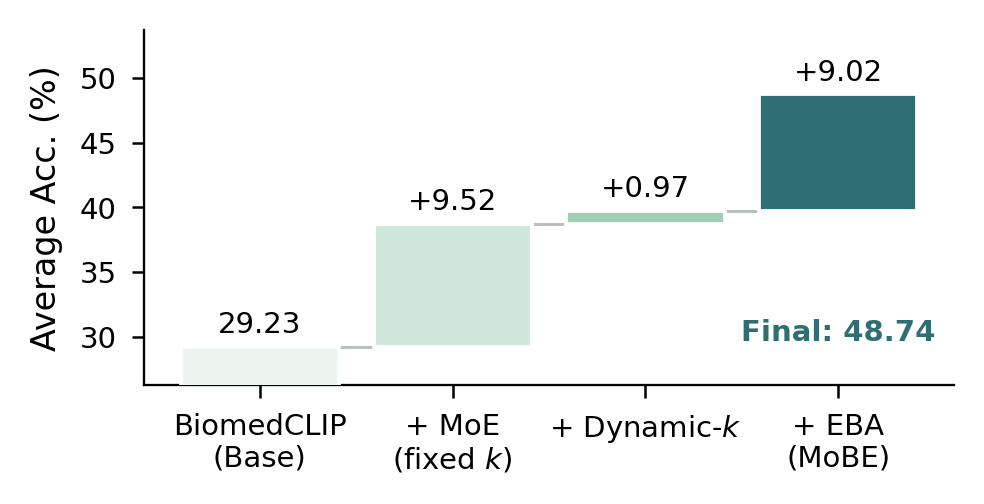}
        \caption{\small Component-wise result}
        \label{fig:compnent_ablation}
    \end{subfigure}
    \hfill
    \begin{subfigure}[t]{0.32\linewidth}
        \centering
        \includegraphics[width=\linewidth]{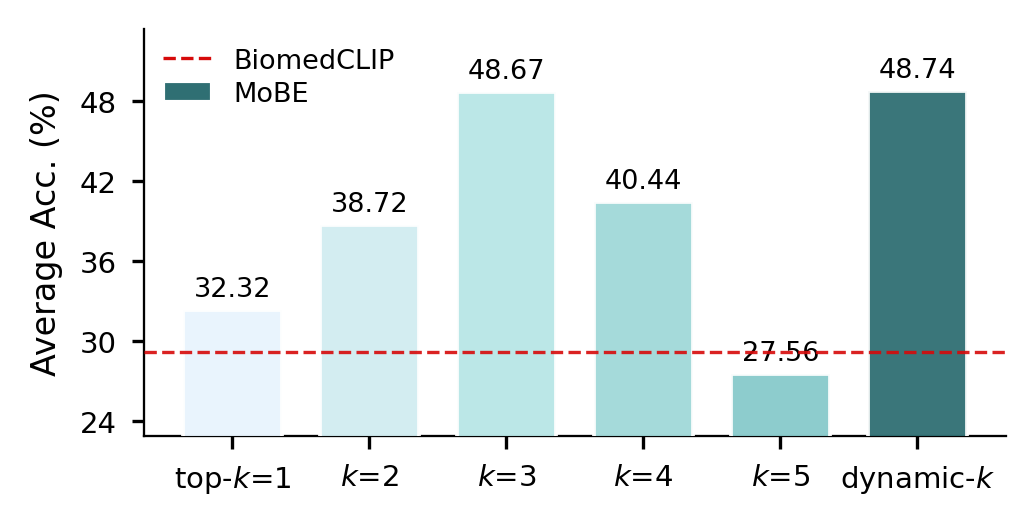}
        \caption{\small Fixed-$k$ \textit{vs.} dynamic-$k$}
        \label{fig:topk_ablation}
    \end{subfigure}
    \begin{subfigure}[t]{0.32\linewidth}
        \centering
        \includegraphics[width=\linewidth]{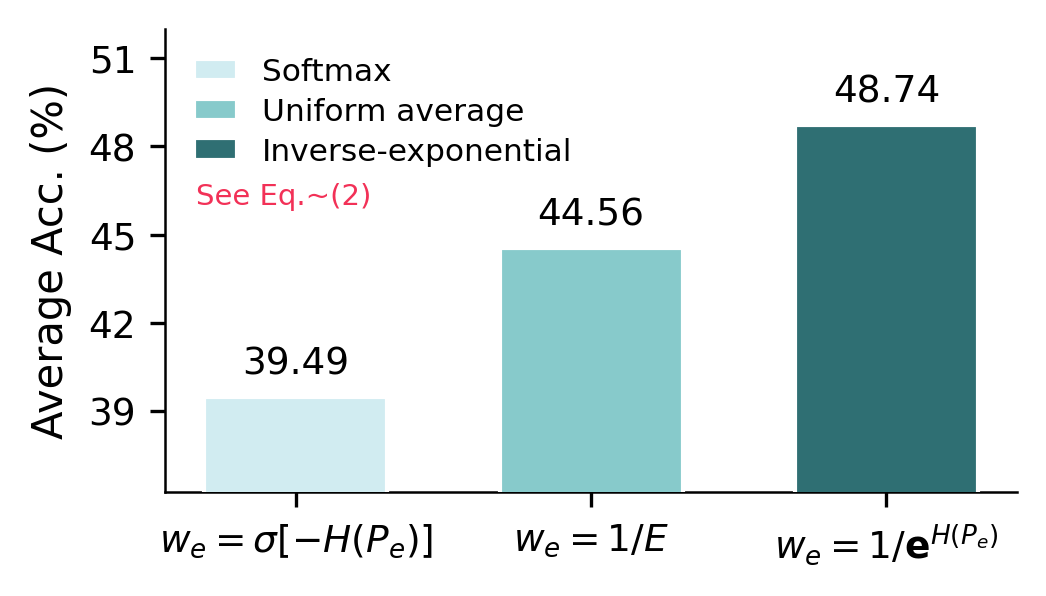}
        \caption{\small Expert weighting effect}
        \label{fig:gating}
    \end{subfigure}
    \caption{
    \small 
        (a) \textbf{Component-wise performance:} Each component of MoBE improves generalization. 
        (b) \textbf{Expert selection effect:} MoBE's dynamic-$k$ routing matches the best fixed top-$k$ while remaining fully automatic. 
        (c) \textbf{Expert weighting effect:} MoBE's inverse-exponential entropy outperforms averaging and softmax as gating functions.
    }
    \label{fig:ablation_3}
\end{figure}

\subsection{Ablation Studies}
We ablate MoBE components by averaging performance over \textit{seen} modalities (ChestMNIST, OrganCMNIST) and \textit{unseen} modalities (PathMNIST, DermaMNIST), assessing the impact of different design choices.

\noindent\textbf{Component-wise Ablation:}
Relative to frozen BiomedCLIP, MoE with fixed top-$k$ improves accuracy; dynamic-$k$ further improves while eliminating the need to tune $k$. EBA yields the largest gain via online refinement of per-expert posteriors, and combined routing-and-adaptation performs best (Fig.~\ref{fig:compnent_ablation}).

\noindent\textbf{Fixed Top-$k$ vs.\ Adaptive-$k$:}
Fixed top-$k$ is sensitive to $k$ (too many experts can hurt). Dynamic-$k$ matches the best fixed-$k$ performance while remaining fully automatic, expanding the expert set only when additional experts have comparable uncertainty (Fig.~\ref{fig:topk_ablation}).

\noindent\textbf{Expert Weighting Mechanism:}
For gating the selected experts, uniform averaging improves over softmax weighting, while inverse-exponential entropy weighting performs best, emphasizing low-entropy experts (Fig.~\ref{fig:gating}).

\noindent\textbf{Compute Analysis:}
Compared to optimization-based MoME, MoBE is inference-only (no backprop), achieving higher accuracy with substantially lower overhead. The cost comes from extra forward passes over multiple experts, making it slower than single-model BiomedCLIP inference (Tab.~\ref{tab:compute}).

\noindent\textbf{Effect of Different Backbones:}
Across backbones (e.g., PubmedCLIP vs.\ BiomedCLIP), MoBE consistently improves average accuracy, with larger gains on unseen modalities, indicating robust backbone transferability (Tab.~\ref{tab:backbone}).

\section{Conclusion}
We address test-time modality generalization in MVLMs with our proposed MoBE: a test-time route-and-adapt MoE framework combining adaptive expert selection and Bayesian adaptation.
MoBE consistently improves performance across seen, unseen, and heterogeneous benchmarks across backbones. However, MoBE incurs higher inference cost than frozen BiomedCLIP due to multiple expert forward passes. Future work should reduce this overhead (e.g., via early-exit routing, pruning/distillation, or caching) and improve robustness when available expert-bank coverage of the test distribution is limited.

%
%
%

%
%
%

\subsubsection*{Disclosure of Interests.} The authors have no competing interests to declare that are relevant to the content of this article.

\bibliographystyle{splncs04}
\bibliography{mybib}

@inproceedings{wang2025test,
  title={Test-Time Adaptation of Medical Vision-Language Models with Mixture of Modality Experts},
  author={Wang, Hancong and Yu, Yue and Zheng, Hairong and Zhang, Tong},
  booktitle={Proceedings of the 33rd ACM International Conference on Multimedia},
  pages={4649--4658},
  year={2025}
}

@inproceedings{karmanov2024efficient,
  title={Efficient test-time adaptation of vision-language models},
  author={Karmanov, Adilbek and Guan, Dayan and Lu, Shijian and El Saddik, Abdulmotaleb and Xing, Eric},
  booktitle={Proceedings of the IEEE/CVF Conference on Computer Vision and Pattern Recognition},
  pages={14162--14171},
  year={2024}
}

@inproceedings{zhou2025bayesian,
  title={Bayesian test-time adaptation for vision-language models},
  author={Zhou, Lihua and Ye, Mao and Li, Shuaifeng and Li, Nianxin and Zhu, Xiatian and Deng, Lei and Liu, Hongbin and Lei, Zhen},
  booktitle={Proceedings of the Computer Vision and Pattern Recognition Conference},
  pages={29999--30009},
  year={2025}
}

@article{han2024dota,
  title={Dota: Distributional test-time adaptation of vision-language models},
  author={Han, Zongbo and Yang, Jialong and Wang, Guangyu and Li, Junfan and Xu, Qianli and Shou, Mike Zheng and Zhang, Changqing},
  journal={arXiv preprint arXiv:2409.19375},
  year={2024}
}

@inproceedings{imam2025robustness,
  title={On the robustness of medical vision-language models: Are they truly generalizable?},
  author={Imam, Raza and Marew, Rufael and Yaqub, Mohammad},
  booktitle={Annual Conference on Medical Image Understanding and Analysis},
  pages={233--256},
  year={2025},
  organization={Springer}
}

@inproceedings{imam2025noise,
  title={Noise is an efficient learner for zero-shot vision-language models},
  author={Imam, Raza and Hanif, Asif and Zhang, Jian and Dawoud, Khaled Waleed and Kementchedjhieva, Yova and Yaqub, Mohammad},
  booktitle={Proceedings of the IEEE/CVF International Conference on Computer Vision},
  pages={5820--5829},
  year={2025}
}

@inproceedings{imam2025test,
  title={Test-time low rank adaptation via confidence maximization for zero-shot generalization of vision-language models},
  author={Imam, Raza and Gani, Hanan and Huzaifa, Muhammad and Nandakumar, Karthik},
  booktitle={2025 IEEE/CVF Winter Conference on Applications of Computer Vision (WACV)},
  pages={5449--5459},
  year={2025},
  organization={IEEE}
}

@inproceedings{ju2025delving,
  title={Delving into out-of-distribution detection with medical vision-language models},
  author={Ju, Lie and Zhou, Sijin and Zhou, Yukun and Lu, Huimin and Zhu, Zhuoting and Keane, Pearse A and Ge, Zongyuan},
  booktitle={International Conference on Medical Image Computing and Computer-Assisted Intervention},
  pages={133--143},
  year={2025},
  organization={Springer}
}

@article{korevaar2023failure,
  title={Failure to achieve domain invariance with domain generalization algorithms: An analysis in medical imaging},
  author={Korevaar, Steven and Tennakoon, Ruwan and Bab-Hadiashar, Alireza},
  journal={IEEE Access},
  volume={11},
  pages={39351--39372},
  year={2023},
  publisher={IEEE}
}

@article{cheng2025understanding,
  title={Understanding the robustness of vision-language models to medical image artefacts},
  author={Cheng, Zijie and Ong, Ariel Yuhan and Wagner, Siegfried K and Merle, David A and Ju, Lie and Zhang, Hanyuan and Chen, Ruinian and Pang, Linze and Li, Boxuan and He, Tiantian and others},
  journal={NPJ Digital Medicine},
  volume={8},
  number={1},
  pages={727},
  year={2025},
  publisher={Nature Publishing Group UK London}
}

@article{shu2022test,
  title={Test-time prompt tuning for zero-shot generalization in vision-language models},
  author={Shu, Manli and Nie, Weili and Huang, De-An and Yu, Zhiding and Goldstein, Tom and Anandkumar, Anima and Xiao, Chaowei},
  journal={Advances in Neural Information Processing Systems},
  volume={35},
  pages={14274--14289},
  year={2022}
}

@article{zhang2024boostadapter,
  title={Boostadapter: Improving vision-language test-time adaptation via regional bootstrapping},
  author={Zhang, Taolin and Wang, Jinpeng and Guo, Hang and Dai, Tao and Chen, Bin and Xia, Shu-Tao},
  journal={Advances in Neural Information Processing Systems},
  volume={37},
  pages={67795--67825},
  year={2024}
}

@article{zhong2022meta,
  title={Meta-dmoe: Adapting to domain shift by meta-distillation from mixture-of-experts},
  author={Zhong, Tao and Chi, Zhixiang and Gu, Li and Wang, Yang and Yu, Yuanhao and Tang, Jin},
  journal={Advances in Neural Information Processing Systems},
  volume={35},
  pages={22243--22257},
  year={2022}
}

@article{li2022sparse,
  title={Sparse mixture-of-experts are domain generalizable learners},
  author={Li, Bo and Shen, Yifei and Yang, Jingkang and Wang, Yezhen and Ren, Jiawei and Che, Tong and Zhang, Jun and Liu, Ziwei},
  journal={arXiv preprint arXiv:2206.04046},
  year={2022}
}

@article{su2026variational,
  title={Variational Inference, Entropy, and Orthogonality: A Unified Theory of Mixture-of-Experts},
  author={Su, Ye and Liu, Yong},
  journal={arXiv preprint arXiv:2601.03577},
  year={2026}
}

@article{nikolic2025exploring,
  title={Exploring expert specialization through unsupervised training in sparse mixture of experts},
  author={Nikolic, Strahinja and Oguz, Ilker and Psaltis, Demetri},
  journal={arXiv preprint arXiv:2509.10025},
  year={2025}
}

@article{mo2024large,
  title={A large-scale medical visual task adaptation benchmark},
  author={Mo, Shentong and Luo, Xufang and Wang, Yansen and Li, Dongsheng},
  journal={arXiv preprint arXiv:2404.12876},
  year={2024}
}

@inproceedings{koleilat2025biomedcoop,
  title={Biomedcoop: Learning to prompt for biomedical vision-language models},
  author={Koleilat, Taha and Asgariandehkordi, Hojat and Rivaz, Hassan and Xiao, Yiming},
  booktitle={Proceedings of the Computer Vision and Pattern Recognition Conference},
  pages={14766--14776},
  year={2025}
}

@article{ruckert2024rocov2,
  title={Rocov2: Radiology objects in context version 2, an updated multimodal image dataset},
  author={R{\"u}ckert, Johannes and Bloch, Louise and Br{\"u}ngel, Raphael and Idrissi-Yaghir, Ahmad and Sch{\"a}fer, Henning and Schmidt, Cynthia S and Koitka, Sven and Pelka, Obioma and Abacha, Asma Ben and G. Seco de Herrera, Alba and others},
  journal={Scientific Data},
  volume={11},
  number={1},
  pages={688},
  year={2024},
  publisher={Nature Publishing Group UK London}
}

@article{zhang2023biomedclip,
  title={Biomedclip: a multimodal biomedical foundation model pretrained from fifteen million scientific image-text pairs},
  author={Zhang, Sheng and Xu, Yanbo and Usuyama, Naoto and Xu, Hanwen and Bagga, Jaspreet and Tinn, Robert and Preston, Sam and Rao, Rajesh and Wei, Mu and Valluri, Naveen and others},
  journal={arXiv preprint arXiv:2303.00915},
  year={2023}
}

@article{wang2020tent,
  title={Tent: Fully test-time adaptation by entropy minimization},
  author={Wang, Dequan and Shelhamer, Evan and Liu, Shaoteng and Olshausen, Bruno and Darrell, Trevor},
  journal={arXiv preprint arXiv:2006.10726},
  year={2020}
}

@inproceedings{niu2022efficient,
  title={Efficient test-time model adaptation without forgetting},
  author={Niu, Shuaicheng and Wu, Jiaxiang and Zhang, Yifan and Chen, Yaofo and Zheng, Shijian and Zhao, Peilin and Tan, Mingkui},
  booktitle={International conference on machine learning},
  pages={16888--16905},
  year={2022},
  organization={PMLR}
}

@inproceedings{zhou2022conditional,
  title={Conditional prompt learning for vision-language models},
  author={Zhou, Kaiyang and Yang, Jingkang and Loy, Chen Change and Liu, Ziwei},
  booktitle={Proceedings of the IEEE/CVF conference on computer vision and pattern recognition},
  pages={16816--16825},
  year={2022}
}

@article{zhou2022learning,
  title={Learning to prompt for vision-language models},
  author={Zhou, Kaiyang and Yang, Jingkang and Loy, Chen Change and Liu, Ziwei},
  journal={International journal of computer vision},
  volume={130},
  number={9},
  pages={2337--2348},
  year={2022},
  publisher={Springer}
}

@article{osowiechi2024watt,
  title={Watt: Weight average test time adaptation of clip},
  author={Osowiechi, David and Noori, Mehrdad and Vargas Hakim, Gustavo and Yazdanpanah, Moslem and Bahri, Ali and Cheraghalikhani, Milad and Dastani, Sahar and Beizaee, Farzad and Ayed, Ismail and Desrosiers, Christian},
  journal={Advances in neural information processing systems},
  volume={37},
  pages={48015--48044},
  year={2024}
}

@article{lee2024entropy,
  title={Entropy is not enough for test-time adaptation: From the perspective of disentangled factors},
  author={Lee, Jonghyun and Jung, Dahuin and Lee, Saehyung and Park, Junsung and Shin, Juhyeon and Hwang, Uiwon and Yoon, Sungroh},
  journal={arXiv preprint arXiv:2403.07366},
  year={2024}
}

@inproceedings{wang2025moe,
  title={MoE-Health: A Mixture of Experts Framework for Robust Multimodal Healthcare Prediction},
  author={Wang, Xiaoyang and Yang, Christopher},
  booktitle={Proceedings of the 16th ACM International Conference on Bioinformatics, Computational Biology, and Health Informatics},
  pages={1--9},
  year={2025}
}

@inproceedings{hanif2024baple,
  title={Baple: Backdoor attacks on medical foundational models using prompt learning},
  author={Hanif, Asif and Shamshad, Fahad and Awais, Muhammad and Naseer, Muzammal and Khan, Fahad Shahbaz and Nandakumar, Karthik and Khan, Salman and Anwer, Rao Muhammad},
  booktitle={International Conference on Medical Image Computing and Computer-Assisted Intervention},
  pages={443--453},
  year={2024},
  organization={Springer}
}

@inproceedings{khan2024guardian,
  title={Guardian: Guarding against uncertainty and adversarial risks in robot-assisted surgeries},
  author={Khan, Ufaq and Nawaz, Umair and Sheikh, Tooba T and Hanif, Asif and Yaqub, Mohammad},
  booktitle={International Workshop on Uncertainty for Safe Utilization of Machine Learning in Medical Imaging},
  pages={59--69},
  year={2024},
  organization={Springer}
}
\end{document}